\documentclass{article}

\usepackage{xcolor}

\usepackage{amsmath}
\usepackage{microtype}
\usepackage{graphicx}
\usepackage{subfigure}
\usepackage{booktabs}
\usepackage{rotating}
\usepackage{afterpage}
\usepackage{array}
\usepackage{caption}

\usepackage{amssymb}
\usepackage{pifont}
\newcommand{\xmark}{\ding{55}}%
\usepackage{hyperref}

\usepackage[accepted,nohyperref]{icml2018}

\icmltitlerunning{Table-Based Neural Units: Fully Quantized, Multiply-Free Networks}

\newcommand{\floatspace}{\vspace*{-0.2in}}

\begin{document}

\twocolumn[
  \icmltitle{Table-Based Neural Units:\\Fully Quantizing Networks for Multiply-Free Inference}

\begin{icmlauthorlist}
\icmlauthor{Michele Covell}{goog}
\icmlauthor{David Marwood}{goog}
\icmlauthor{Shumeet Baluja}{goog}
\icmlauthor{Nick Johnston}{goog}
\end{icmlauthorlist}

\icmlaffiliation{goog}{Google Research, Moutain View, CA, USA}
\icmlcorrespondingauthor{Michele Covell}{covell@google.com}

\icmlkeywords{Machine Learning, Quantization, Deep Learning, Deep Architectures, ASIC, FPGA}

\vskip 0.3in
]

\printAffiliationsAndNotice{}

\begin{abstract}

  In this work, we propose to quantize all parts of standard
  classification networks and replace the activation-weight--multiply step
  with a simple table-based lookup.  This approach results in networks that
  are free of floating-point operations and free of multiplications,
  suitable for direct FPGA and ASIC implementations.
  It also provides us with two
  simple measures of per-layer and network-wide compactness
  as well as insight into the distribution
  characteristics of activation-output and weight values.
  We run
  controlled studies across different quantization schemes, both fixed
  and adaptive and, within the set of adaptive approaches, both
  parametric and model-free.
  We implement our approach to
  quantization with minimal, localized changes to the training
  process, allowing us to benefit from advances in training continuous-valued
  network architectures.  We apply our approach successfully to
  AlexNet, ResNet, and MobileNet.  We show results that are within
  1.6\% of the reported, non-quantized performance on MobileNet using
  only 40 entries in our table.  This performance gap narrows to zero
  when we allow tables with 320 entries.  Our results give the best
  accuracies among multiply-free networks.
  
\end{abstract}

\section{Introduction to Network Quantization}
\label{sec:intro}

Deep neural networks are being employed in an ever-growing number
of applications.  Though these networks are often trained on computers
with powerful GPUs, inference is increasingly happening on end-user devices.
To address the enormous range of memory, bandwidth, and computation
constraints that must be handled for successful wide deployment of
neural-network--based applications, there has
been renewed interest in efficient
inference.  One promising technique is network quantization.
Research into neural-network quantization is not new; it has spanned
decades and included numerous
approaches~\cite{Balzer1991,Marchesi1993,yi2008new,Vanhoucke2011,Anwar2015,courbariaux2016,nvidia2016,Hubara2016,Zhou2016,Garland2017,deng2017,wu2018training,Baluja2018b}.
\cite{Guo2018} provides an excellent survey of the prior work in this
area.
His analysis includes
a useful characterization of the methods as using either fixed or
adaptive codebooks
for the
weight quantization.  In this paper, we introduce new adaptive
(Section~\ref{sec:model-free}) and new fixed
(Section~\ref{sec:octave}) quantization approaches.

There is
little agreement on what it means to quantize a network.  For
example,~\cite{courbariaux2016,Li2017} quantize only the
weights but not the biases or the activations,
while \cite{Cai2017} focuses on activations.
Recent efforts, including~\cite{Zhou2016,
  deng2017}, quantized \emph{some} of the weights and \emph{some} of the activations
but not all of them.
Some studies quantize the network uniformly to allow operations using
fixed-point arithmetic~\cite{Jacob2018}, others quantize each channel of
each layer independently~\cite{Krishnamoorthi2018}.
Successful previous studies have employed varying levels of
discretization, ranging from bi-level and tri-level weights to 4- and
8-bit instantiations.
Some use adaptive quantization~\cite{Gong2014,Choi2017,Achterhold2018}
and others use fixed
codebooks~\cite{Tang1993,Hwang2014,Kim2014,Courbariaux2015,Zhu2016}.
Finally, as expected, the degree of change needed in the training algorithms
varies significantly~\cite{rastegari2016, Zhou2016, Tang2017, Zhou2017}.

This range of definitions for what it means to quantize a network
makes it difficult to compare approaches.  In this
study, as in~\cite{Baluja2018b}, we start with the premise that we
should be able quantize \emph{all} parts of the network: the biases,
the weights, and the activations.
Full quantization allows us to avoid all floating-point operations and
even fixed-point multiplication, replacing it with a Look-Up-Table
(LUT).  For an FPGA implementation, there is a 
reasonably direct trade-off between the expense of floating point and
the memory needed for our LUTs: if the LUT size is less than a few thousand
entries, it takes fewer FPGA resources than a CLB-based floating-point
multiply~\cite{Marcus2004}.

Full quantization also allows us to
develop two usable measures with which to compare
quantizations, {\em neural-unit complexity (NUC)} and {\em network-wide non-compactness (NWNC)}.  Inspired
by~\cite{Chatterjee2018}, which used 
LUTs in the study of network generalization and
capacity, we reformulate the standard multiplication
of activations and
weights as a simple table-cell
look-up.  The sizes of the required LUTs give a
partial measure of complexity and compactness: with NUC, we use the
per-layer LUT size as a local measure of complexity and, with NWNC, we
use the combined LUT sizes across the full network as a measure of the
(non-)compactness of the full representation.
With these measures, we are able to compare
adaptive codebooks to
fixed ones, to examine the best allocation between the
activation and
weight quantization
levels, and to examine the gain in NWNC of using a single global set of weight
quantizations compared to layer-specific weight quantizations.

To avoid moving unnecessarily far from the underlying floating-point
networks that we are analyzing, we propose two simple, highly
localized changes to standard deep-learning training regimes.  Though simple, these yield
fully discretized networks with minimal accuracy loss; these are described in 
Section~\ref{sec:train}.
In Section~\ref{sec:weight quant}, we examine four alternative
weight-quantization approaches, using AlexNet~\cite{krizhevsky2012imagenet}
as our target.
In Section~\ref{sec:act quant}, we examine two approaches
to activation quantization: uniform linear and \emph{octave
  sampling}. Using octave sampling for both weights and activations,
we are able to move away from doubly indexed LUTs to much 
smaller single-index tables.
In Section~\ref{sec:results}, we conduct extensive empirical tests, comparing our results to others
on three different-sized networks: AlexNet, for comparison
with~\cite{wu2018training, Zhou2016, Hubara2016,
  rastegari2016, qualcomm2017}); ResNet-101, for a large-network study; and
MobileNet, for networks targeted at mobile devices.  Finally,
conclusions are presented in Section~\ref{sec:conclude}.

\section{LUT-based Networks}
\label{sec:train}

In this section, we present  the quantization procedure,
originally described in~\cite{Baluja2018b}.
We quantize every portion of the network: all weights (including
biases) and activations. For ease in exposition, we begin with
the process of inference used in our quantized LUT-based
networks.
Then we return to training.

\subsection{Inference using LUT-based Networks}
\label{sec:inference}

\begin{figure}[t]
  \includegraphics[width=\linewidth]{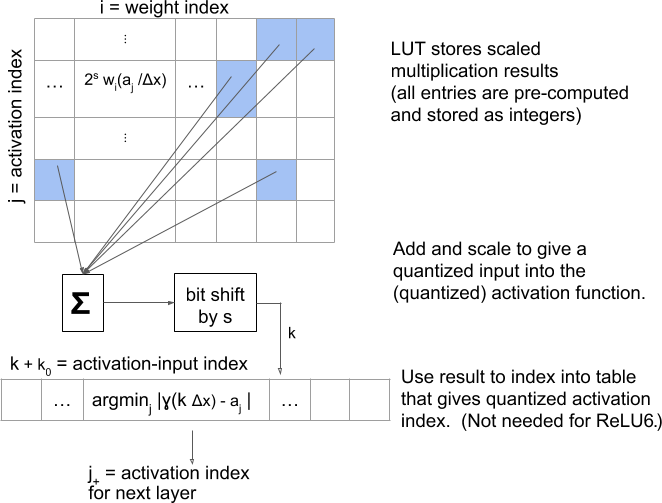}
  \floatspace
  
  \caption{\footnotesize LUT-based neural unit. $\text{LUT}(i, j) =
    \frac{2^s}{\Delta x} w_i a_j$ (for
    quantized weight
    $w_i$ and quantized activation-output $a_j$)
    replaces
    \emph{all} multiplies at
    inference time.  These values are summed to give
    $2^s\frac{x}{\Delta x}$ where $x$ is the output of the linear unit.
     and the input into the next non-linearity $\Gamma()$.
    Shifting by $s$ bits removes the $2^s$
    scaling.  We then use the activation table, indexed by
    $k + k_0$ where $k = \frac{x}{\Delta x}$ (the output from our
    sum/shift) and $k_0$ = index for $x=0$.  The
    values in the activation table are the index $j_+$ for the
    quantized activation, $a_{j_+} = \arg \min_j | \Gamma(k\Delta
    x) - a_j |$. If there is no
    following layer, we can look up that actual value of $\Gamma(k \Delta
    x)$ by looking up its value at in the LUT, using the value of $i$
    that corresponds to $w_i = 1$.
    No multiplications and no non-linearities are computed at
    inference time and all additions are integer.
    Note that the number
    of entries in the activation table can be more than $N_a$ (the
    number of distinct quantized activation levels) if our nonlinearity
    does not change level at a uniform rate (e.g., quantized tanh).
    {\em (Similar to Figure~9 in~\cite{Baluja2018b}, with permission.)}
    }
  \label{fig:methodNoMult}
  \floatspace
\end{figure}

To eliminate {\em all} multiplications and {\em all} floating-point
operations,~\cite{Baluja2018b} used doubly-indexed LUTs that hold the
product of all the
quantized weights and activations.\footnote{
  Note that this approach does not require uniform quantization on either
  weights or activations.}
Figure~\ref{fig:methodNoMult} illustrates this approach.
After quantization of weights and activations, the
network is represented with three tables; Figure~\ref{fig:methodNoMult} shows two of
them. The third table, the weight-index table, is also discussed below.

The first table, the LUT, provides
the result of the multiplication between a quantized
weight and its input, which is, itself, the output of the previous
layer's quantized activation function.  This table contains $N_a N_w$
elements: the number of activation quantization levels, 
$N_a$, times the number of weight quantization levels, $N_w$. The value 
stored in each cell is the closest integer to $\frac{2^s}{\Delta x}
w_i a_j$ where $2^s$
is the fixed-point scaling factor
(allowing accurate accumulation) and $\Delta x$ is the
quantization step used in the activation table.  We force 
the weight quantizations to include one;
this also handles the bias elements.
Additionally, the  $w_i=1$ column can be used to look-up the final layer's
actual output value.\footnote{Similarly, to implement the scaling
  needed for average pooling, we can add a single-row for that
  specific scaling.}

This LUT is different at each layer, if the quantization levels for
the different layers are not shared.  If the network computation is
scheduled by layer,
then only the current layer's LUT needs to fit into CPU L1 cache
(or the FPGA block RAM) at a
time, but there are time and energy costs to updating the
tables between layers.  We try to capture these two types of cost by
breaking out the per-layer LUT size with NUC and the full network cost
using NWNC.

The second table, the activation table, is of size $N_x$, where $N_x$
is the number of $\Delta x$-sized steps needed in the activation-input
space to fully span the quantized range of activation outputs.  Assuming a
standard,
bounded, non-decreasing
activation function, $\gamma( )$
(e.g., Tanh, Sigmoid, ReLu6), $N_x$ is the
count from the largest $k$ giving $\arg \min_j | \gamma(k
\Delta x) - a_j | = 0$ through (inclusive) the smallest $k$ giving
$\arg \min_j | \gamma(k \Delta x) - a_j | = N_a - 1$.  For example, if
$\gamma( )$ is Tanh, quantized to uniform output spacing $\frac{2}{N_a
  - 1}$ with $\Delta x = 0.02$ and $N_a = 32$, then $N_x = 207$
(across $x = \pm 2.06$).
It is important to note here that the size of the LUT and the
activation table are crucial for memory and bandwidth constrained
systems.

With an FPGA implementation, ideally, both the LUT and activation
table
should simultaneously fit
onto the distributed memory of the FPGA,
since the order in which they will be accessed is not
predictable.  This is easier than the full LUT size would suggest,
since only the one column of the LUT that corresponds to $w_i$ (the
weight that is used at that location in the network) needs to be
accessed during each multiplication.  For the purposes of FPGA access
(where the processing and memory is distributed), this reduces the
per-multiplication memory usage to $N_a$ and the per-accumulator
memory usage to $N_x$.

For the final classification decision, we retain the full precision
of the accumulator instead of completing the ``$\ll$ s'',
indicated in Figure~\ref{fig:methodNoMult}.  We then find the index
of the 1$^{st}$ (for recall@1) or top-5 (for recall@5) maximum output
values.  In this way, we do not lose ranking precision and we do not
need to move to floating-point numbers.

The third table is the weight-index. For each of the network's
connections, this stores the quantized-weight index.
This is the largest table; it has
$N_{\text{net}}$ entries, where $N_{\text{net}}$ is the
number of weights and biases within the network.  Each of the
$N_{\text{net}}$ entries
is $\left\lceil \log_2 N_w
\right\rceil$ bits wide,
(where $N_w$ is the \# of weight quantization levels).\footnote{
  The average size of the weight-index--table entries
  ($\left\lceil \log_2 N_w \right\rceil$) can be further reduced using
  variable-length encoding.  However, since this study is focused on
  the on-chip table
  sizes and operations, we do not attempt to quantify the amount of additional
  compression available by that route.}
Fortunately, since the sequencing for the neural-network computation
is both fixed and predictable, this table can be accessed  in
predefined sections efficiently.  It does not need to fit directly on
chip. 
Nonetheless, limiting the table size has secondary benefits, 
since it 
impacts the number of total table accesses and network download size.

A note about network inputs: we handle the network inputs into the
first layer either by quantizing the input values to the network's
activation quantization levels or by using a separate LUT for
multiplying the network's inputs by the same
quantized weights as used elsewhere.  Looking ahead, when we used the
former approach,
we saw a drop in accuracy of 0.0--0.3\% from what is reported in
Table~\ref{table: compare}.  \cite{Baluja2018b} provides more details about
the impact of input quantization on k-means--based weight quantization
(Section~\ref{sec:kmeans}) and Laplacian-model--based weight quantization
(Section~\ref{sec:laplacian})

In summary, using the standard floating-point representations for
comparison, our approach reduces the size of the overall
network from $32 N_{\text{net}}$ down to $N_{\text{net}} \log_2 N_w$
 (for the weight table) $+$ $(s + \log_2 N_x)
N_a N_w$ (for the LUT) $+$ $N_x \log_2 N_a$ (for the activation
table).  Since the values of $N_w$, $N_a$, and $N_x$ are
miniscule compared to $N_{\text{net}}$, there is an immediate
reduction in network download/storage size by
$\frac{32}{\lceil\log_2 N_w\rceil}$.
Additionally, as we mentioned above,  the LUT sizes provide intuitive
measures with which to compare different quantization allocations and schemes.

\subsection{Training Fully-Quantized Networks}
\label{sec:training}

To quantize the entire inference network, we broadly define two sets of
quantization levels.  The first set is used for the weights
and bias units.
The second set of levels is used for the
activations.  In its simplest form, the product of these two
 sets
determines the NUC since it represents the total set of 
unique values that can be produced at any single layer or unit.   Our
goal is to 
minimize the set sizes of both, while also minimizing the impact to
classification accuracy.  To do this, we change the training process
in two simple, localized steps, in the same way as was done
in~\cite{Baluja2018b}.

The first modification allows backwards error-propagation
through quantized activations.
We use the ``straight-through estimator''
(STE)~\cite{hinton2012lecture}, which has been widely used
for quantization~\cite{Courbariaux2015, Zhou2016, Miyashita2016}.
Simply, this means that during training, the
discretized activations use the gradients that would have
been provided by the non-discretized version.

The second training modification allows us to find good weight and bias
cluster centers.
Using one of the methods from Section~\ref{sec:weight
  quant}, we find new quantization centers once every $S$ training
steps~\footnote{Note that for the methods described in
  Section~\ref{sec:model-free} and ~\ref{sec:octave}, the 
  {\em cluster centers} will not change after the
  first quantization pass.}
(we set $S=1000$).  In each of these quantizing passes, we replace each weight
(and bias) with its assigned quantization center.  This temporarily
reduces the number of unique weights and biases to $N_w$.  After this
quantization event, training continues with no modifications until
the next multiple-of-$S$ training step.  Between the quantization
events, the weights diverge from the quantized levels.  Nonetheless,
these periodic reinforcements of the selected levels ensures that the
final quantization event is not detrimental to performance.
More details can be found in~\cite{Baluja2018b}.

\subsection{Handling Batch-Norm for Quantization}
\label{sec:norm}

One hurdle to quantization is batch-norm (BN)~\cite{Ioffe2015}.
The difficulty for quantization arises because BN moves the
weights and biases of each unit independently.  Most earlier
quantization work has
focused either on networks that are shallow enough that they can be
trained without using BN or on quantization approaches that rescale
quantization levels across layers~\cite{Jacob2018} and, often, even
across channels~\cite{Krishnamoorthi2018}.
Unfortunately, the
first (training without BN) limits the set of networks that we
can consider.  The latter (using different quantization scaling)
increases the distinct
quantization levels used by the weights and biases across the full
network, giving much larger values for NWNC.  By folding 
BN into the weights \emph{before quantizing}, we can avoid both of
these compromises.  

We first train the network without
quantization,
using BN.
Upon convergence, we fold the changes dictated by BN
into the
weight layers immediately preceding the normalization functions. We thereby
eliminate the $1\times1\times1$ linear layer that BN would otherwise
create.
The BN changes to the weights and biases are 
combined with those of the basic convolutional unit, using the
equations given in Appendix~\ref{supplemental sec:norm},
and the BN unit can be removed.  
We then 
measure 
the distribution of these ``BN-folded'' weights and biases
(for Section~\ref{sec:model-free}) or the extremal values
(for Section~\ref{sec:octave}).
The BN-folded weights and biases are then quantized using one
of the approaches described next.

\section{Weight quantization approaches}
\label{sec:weight quant}

We present four  approaches to weight/bias
quantization.  In this section,
our activation
quantization
is uniform.
For the purposes of being
concrete,
we compare these alternative
quantization approaches using AlexNet~\cite{krizhevsky2012imagenet}.
Our baseline accuracy with AlexNet (without
quantization, using ReLU6) on ImageNet~\cite{deng2009imagenet} is
56.4\% / 79.8\% (recall@1 / recall@5).
More training and baseline-accuracy details will be
given in Section~\ref{sec:results}.

\subsection{Weight/bias quantization using 1-D k-means}
\label{sec:kmeans}

Previous efforts in adaptive selection of weight quantizations have
used k-means~\cite{jain2010data} on the weight vectors that make up
the full kernels~\cite{Gong2014, Han2015, Choi2017}.
\cite{Baluja2018b} took a
simpler approach here and performed k-means clustering on weights
and biases: in effect, one-dimensional clustering.
Though the optimal solution can be found in $O(n^2 k)$~\cite{wang2011ckmeans}, because of the size of the networks considered ($> 5*10^7$ for AlexNet),
\cite{Baluja2018b} instead used the standard k-means algorithm 
and subsampled the weight/biases to $n=100,000$.

Using this approach with AlexNet, starting quantization training from scratch, using
1000 weight quantization levels (across \emph{all} weights and biases in the
network), and 32 linearly spaced activation levels with
ReLU6, the classification accuracy is 52.5\% (recall@1) and
76.3\% (recall@5): a drop of 3.9\% and 3.5\%, respectively, from our
baseline.
The NUC, the set size of the possible outputs for each
unit, is $32*1000=32,000$ and, since the same levels are used across the full network, the NWNC is also 32,000.

\subsection{Quantization using a Laplacian distribution model}
\label{sec:laplacian}

\begin{figure}
\centering

  \includegraphics[width=0.48\linewidth]{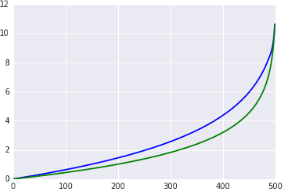}\hfill\includegraphics[width=0.48\linewidth]{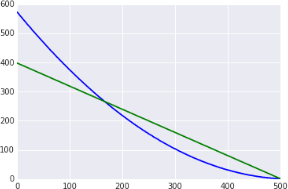}

\caption{\footnotesize Quantization centers (left) and bin counts
  (right), for the upper half
  of a Laplacian distribution $\sigma = \sqrt{2}$,
  $N_w=1000$, $N_{\text{net}}=100,000$.  Green minimizes
  $L_1$ quantization error;
  blue,
  $L_2$ error.
  Quantization centers are non-uniformly spaced with wider
  spacing at large amplitudes.
  {\em (Reproduced with permission from \cite{Baluja2018b})}
}
\label{fig:DistributionModel}
\floatspace

\end{figure}

The k-means approach from Section~\ref{sec:kmeans}
used subsampling to give a non-parametric
estimate of the weight distribution.
However, the weight distributions of a fully trained AlexNet
appear to be nearly Laplacian (or, for
some layers, Gaussian).
This insight opens up
novel model-based quantization approaches. 
Specifically, we can mathematically
determine what the correct distribution of cluster centers and cluster
occupancies should be to minimize {\em expected} $L_1$ or $L_2$ error
for a Laplacian.
We find this lowers the total error across the full weight set,
compared to k-means on the 0.2\% sample used in
Section~\ref{sec:kmeans}.
Figure~\ref{fig:DistributionModel} shows these
cluster centers and occupancies for a sample set drawn from a
Laplacian distribution.  More details are provided in
Appendix~\ref{supplemental sec:laplacian}.

As in Section~\ref{sec:kmeans}, we trained AlexNet from scratch and
used 1000 weight/bias quantization levels and 32 linearly spaced ReLU6
activation levels (the same neural-unit complexity of 32,000
separate entries).
We achieved 57.1\% (recall@1) and 79.8\%
(recall@5).  This is slightly {\em better} recall@1 than even the baseline, 
and the same recall@5 as the baseline.  Both of these
are better than  k-means clustering
(Section~\ref{sec:kmeans}), without increasing either NUC or NWNC.

\subsection{Quantization using model-free distributions}
\label{sec:model-free}

If the optimal continuous-valued weights/biases are a sample from a
single Laplacian distribution, we can minimize the expected $L_1$ (or $L_2$)
error between the quantization centers and the optimal values using
either a set of pre-computed cluster centers
(Figure~\ref{fig:DistributionModel}-a) or a set of pre-computed cluster
occupancy counts (Figure~\ref{fig:DistributionModel}-b).
In Section~\ref{sec:laplacian}, we used the pre-computed cluster centers.
Here, we use the
pre-computed cluster-occupancy counts.
Figure~\ref{fig:DistributionModel}-b shows
that the
occupancy counts for minimizing the expected $L_1$ Laplacian error is a
(discretized) symmetric-about-zero triangle whose width is $N_w + 2$
and whose area is $N_{\text{net}}$: there are no scale parameters to be estimated.
We simply collect, for each layer separately, the fully-trained,
continuous-valued set of weights
and biases; sort them;
and assign them to the quantization bins according to the cumulative
counts that are indicated by the integral of the discretized triangle
shape.  The cluster centers are computed as the median
value for each bin (to minimize $L_1$ error in the bin) or as
the mean value (for $L_2$ error).

We are relying on the observed distributions to determine the
location of our cluster centers. So, we begin by first training the model
{\em without discretization}. We then use the distribution of the
trained, continuous-valued weights/biases, along with our
Laplacian-inspired triangle-occupancy profile, to determine the cluster
occupancy count and our cluster centers.  (The complete procedure is
described, with step-by-step pseudo-code, in
Appendix~\ref{supplemental sec:model-free-weight-quant}.)
We freeze both the cluster centers and cluster occupancy
counts for the remainder of the quantized training, to
avoid the problem of the weight/bias distribution straying from those
found during the continuous-valued training.

If we  quantized once and discontinued training, there would be a large
drop in classification accuracy. 
With AlexNet, in the {\em first} quantization pass to 256 levels, the
recall@1 was 10\% below the final accuracy obtained by
fine tuning.
So, we continue training, using this quantized starting point.
As expected, with continued training, the quantized weights  move away
from the quantized values.  Periodically (every 1000$^{th}$ step), we
again sort the biases and weights and reassociate each with the
cluster center that the triangle-occupancy profile dictates,
regardless of whether that is the closest cluster center by $L_1$
distance.  These weights/biases take on the value
of the cluster center --- the value that was determined (and frozen)
based on our initial quantization.
The individual weights/biases can change which cluster level
they are quantized to, but the overall distribution is constant.
Note the  contrast with Sections~\ref{sec:kmeans} and~\ref{sec:laplacian}.:
both of those approaches re-clustered the weights/biases
to whichever quantization level was closest,
regardless of occupancy counts.

This approach allows us to use units with lower complexity than
before with no loss in recall@5 accuracy.  Using 256 quantized
weight/bias levels (instead of the 1000 previously) and
32 linearly spaced ReLU6 activation levels, the NUC is 8192 entries.
However, since we do the clustering independently at each layer, the
NWNC is less compact than our previous approaches: the NWNC is 65,524
entries.
With this, the
model-free--quantized AlexNet achieves 56.4\% (recall@1) and 79.8\%
(recall@5).  The recall@5 performance matches both
the baseline (unquantized) AlexNet and the Laplacian-modeled
quantization --- with $\frac{1}{4}$ of the
NUC.

\subsection{Weight/bias quantization using %
  octave sampling}
\label{sec:octave}

While we were able to reduce the neural-unit complexity
using the model-free approach, we still used a large look-up
table ($256\times32$ entries) and the NWNC actually got worse (by a
factor of 2).  In this section, we present an
alternative approach: we constrain the weight cluster
centers such that we can re-use entries in our table to represent
distinct weights.
The intuition is to use weights that have the following periodicity
every $N_q$ entries:
 in an ordered list of values to represent, every value at
 index $i$ is simply a
 bit-shifted version of the base value at index $i\mod N_q$.
This is possible using values for the weights/biases that are
separated by exact octaves: specifically, 0 and $$\pm
2^{K_{\max}-k} 2^{-n/N_q}$$ for $0 < n \leq N_q$ and $0 \leq k < N_o$
where $K_{\max} = 2^{\lceil \log_2 v_{\max} \rceil }$ and $N_o =
\lceil \log_2 v_{\max} \rceil - \lfloor \log_2 v_{\min}
  \rfloor$, with $v_{\max}$/$v_{\min}$ as the maximum/minimum non-zero
  amplitude values represented.
Pseudo-code for this octave-based weight quantization approach is given in
Appendix~\ref{supplemental sec:octave-weight-quant}.

Using this approach, we can represent $2 N_q N_o + 1$ (the $+1$ is for
the 0 value) distinct quantized weights with a LUT that is only $N_a
N_q$ separate entries.  This is a factor of $2 N_o\times$ smaller than
would be needed for that many unconstrained entries.  While a
reduction in the NUC and NWNC is accurate, since we are
constraining our weight centers,  %
using
just that LUT size ($N_a N_q$ entries) understates the unit's (and the
network's)
complexity.  With just that measure, there is no accounting for the
number of octaves used. %
To address this, we add another term to our neural complexity
With respect to trainable parameters, we have one additional
parameter for each additional octave that we cover.  So, for
octave-based weight quantization (and linear activation quantization),
our neural complexity is the LUT size ($N_q N_a$) plus the additional
octave parameters ($N_o - 1$): $N_q N_a + N_o - 1$.  For our
octave-based approach, we use the same quantization levels throughout
the network, so the NWNC is the same as the NUC ($N_q N_a + N_o - 1$).

We have observed that octave sampling %
is well matched to the light-tailed
distributions of our network weights/biases, especially on the outer
3--4 octaves.
Octave sampling
oversamples the %
central lobe compared to what Laplacian
distributions dictate, but these values do not require
additional LUT memory.\footnote{It might still be worthwhile pruning
  the number of quantization levels for the lower-amplitude octaves,
  since this number does impact the size of the weight-index table.}

With these quantization centers, we can represent the weight index
using a sign bit ($\pm 1$), an octave ($k$), and a within-octave index
($n$).\footnote{Weights that quantize to zero are dropped before
  being saved.}  When retrieving the result of
multiplying a quantized activation ($a$) with the weight, the
LUT only needs to hold results for $a \times 2^{-n/N_q}$ (so $N_a
N_q$ entries).  The answer can then be determined
by bit shifting the saved result by $K_{\max} - k$ bits and possibly
inverting the sign.

We tested this approach with similar %
numbers of quantization levels as %
Section~\ref{sec:model-free}: 15 octaves with 8
weight/bias samples per octave and 32 linearly spaced ReLU6
activation levels.  Although the number of weight/bias
quantization levels are similar to what we used for model-free
quantization (241 vs 256 levels), the LUT size needed
has only 256 entries, in contrast
to the 8,192 entries in the model free approach.
The neural-unit complexity is increased from the basic LUT size by
the 15 octaves that are used, giving a NUC of $32 \times 9 + 15 -
1 = 271$ entries.  Even with
this constrained representation of weights, octave-quantized
AlexNet gives 56.4\% (recall@1) and 79.8\% (recall@5).
This matches the  performance of the model-free
quantization.

This approach is similar
to~\cite{Miyashita2016}, but with more general spacing in the
log-amplitude space. %
Using this approach with $N_{q} = 1$, $N_{o} = 32$, and $N_a = 16$,
\cite{Miyashita2016} reported an AlexNet recall@5 accuracy of
73.6\%.  Using our approach with the same parameters, we
obtain an accuracy of 76.2\%: a 2.6\% improvement.

\begin{figure*}
  \centering
  \begin{small}
  \begin{minipage}[t]{0.32\linewidth}
AlexNet Results (baseline: 56.4\%) \\
\includegraphics[width=\linewidth]{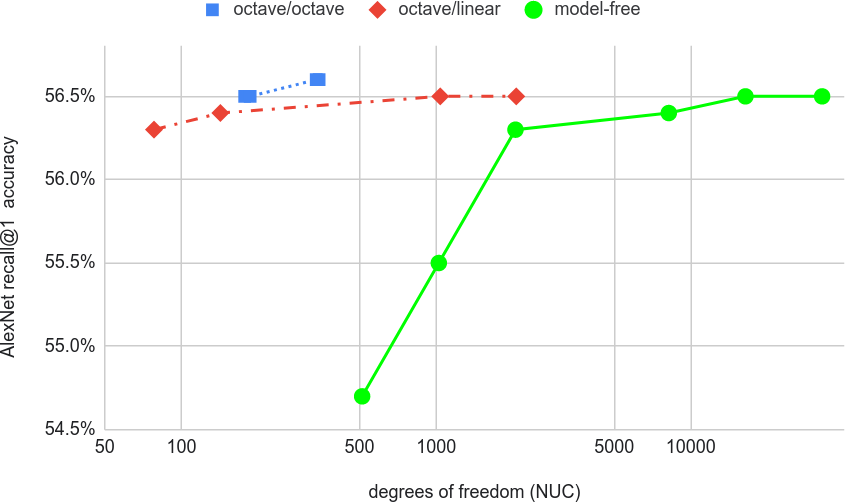}
  \end{minipage}\hfill\begin{minipage}[t]{0.32\linewidth}
ResNet Results (baseline: 74.8\%) \\ 
\includegraphics[width=\linewidth]{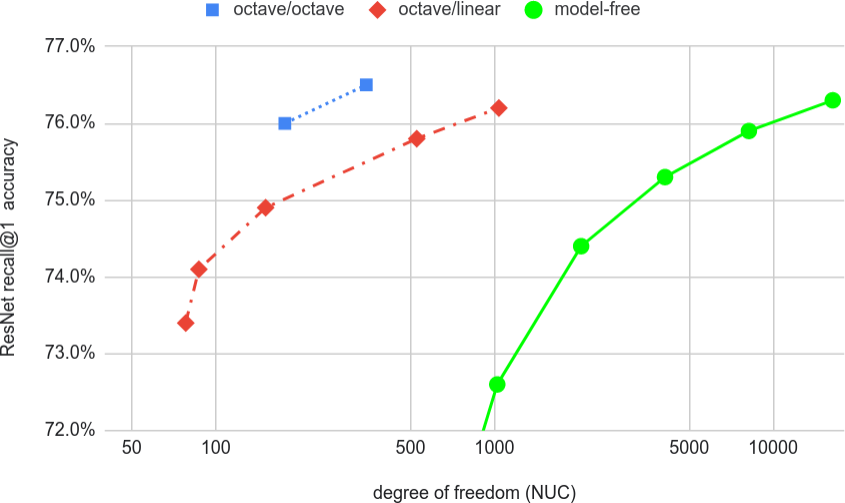}
  \end{minipage}\hfill\begin{minipage}[t]{0.32\linewidth}
MobileNet Results (baseline: 68.5\%) \\
\includegraphics[width=\linewidth]{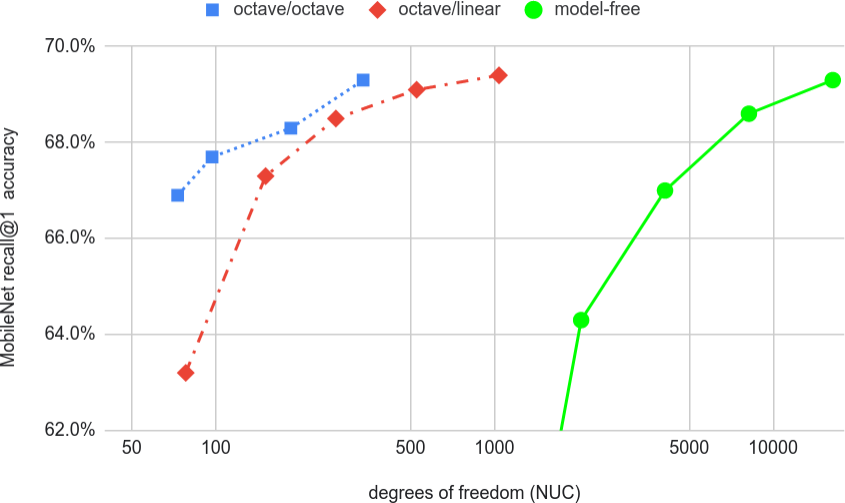}
  \end{minipage}
  \end{small}
\caption{\footnotesize Accuracy versus Neural-Unit Complexity.  The X-axis shows the number of trainable parameters per table-based neural
  unit as a measure of complexity.  Note: the X-axis uses {\em
    log spacing}.  If we had instead used NWNC, ``octave/octave'' and
  ``octave/linear'' would have remained the same but ``model free''
  would have slide to the right by a factor of 8, 101, and 28 for
  AlexNet, ResNet, and MobileNet, respectively.}
\label{fig:AlexNet Results}
\floatspace
\end{figure*} 

\section{Activation Quantization Approaches}
\label{sec:act quant}

In this section, we compare the trade-offs between two alternative
approaches to activation quantization: uniform linear
quantization and octave-based quantization.

\subsection{Uniform-linear activation quantization}

To this point, we have used linearly-spaced quantization levels on our
activations.  Uniform linear quantization is well matched to
the distribution of the activation values, largely due to the
compressive mapping of the bounded activation function.  As %
shown in Section~\ref{sec:octave}, when we use
15 octaves with 8 levels/octaves for weight quantization
with 32-level uniform-linear quantization of activations, we
get 56.4\% (recall@1) and 79.8\% (recall@5) accuracy with AlexNet. %
The NUC and NWNC (both at 271 entries) are the smallest that
we have been able to use successfully, so far. %

\subsection{Octave-based activation quantization}
\label{sec:LUT-free}

By quantizing the weights/biases and activations using log-amplitude quantization, we can
both avoid the multiplication LUT and use only fixed-point addition for
multiplication, by operating in the $\log_2$ index space.

Multiplications between quantized weights and quantized
activations become fixed-point additions.
Multiplying a quantized activation (with $N_{q;a}$ samples per octave)
by a quantized weight (with $N_{q;w}$ samples per octave) is the same
as adding the quantized weight index, scaled by
$\frac{N_{q;a}}{N_{q;w}}$, to the quantized activation index.  As long
as $N_{q;a}$ is a power of two 
times $N_{q;w}$, this combination can be done by bit shifting the
weight index, %
followed by a fixed-point addition.  Importantly, the large
doubly-indexed multiplication LUTs are no longer needed and
are replaced with two, much smaller, single-index tables, as described below.

To sum the weighted inputs into each unit, we move back into (and
then out of) the linear amplitude space using
two small look-up tables. One table, with $\max(N_{q;w}, N_{q;a})$
entries, converts from log to linear amplitudes.  As long as
$\frac{\max(N_{q;w}, N_{q;a})}{\min(N_{q;w}, N_{q;a})}$ is an integer,
we can do this conversion exactly (to the precision of the table
entry). A second table, with $4 N_{q;a}$ entries, returns us into the
log-amplitude space, after accumulation. %
Discussion of the mechanics of this approach to accumulation
starting from log-amplitude and sign representations
are provided in Appendix~\ref{supplemental sec:octave/octave}.

Returning to our definition of NUC, the
complexity of the octave/octave unit would be %
$\max(N_{q;w}, N_{q;a})$ (for the log-to-linear LUT)
plus $4 N_{q;a}$ (for the reverse LUT) plus the numbers of
additional octaves that re-use those LUTs in the weight space
($N_{o;w} - 1$) and in the activation space ($N_{o;a} - 1$).  So the
NUC and NWNC are $\max(N_{q;w}, N_{q;a}) + 4
N_{q;a} + N_{o;w} + N_{o;a} - 2$.

We now revisit AlexNet using 8 weight
samples/octave for 15 octaves. In contrast
with 56.4\% (recall@1) using 32 {\em linearly-spaced} activation levels
(reported earlier), we get only 55.2\% (recall@1) using 96 {\em
  log-spaced} activation levels (specifically, 32
samples/octave for 3 octaves).
This is a loss in performance compared to %
the linearly-quantized activations,
although we actually increased the number of activation
quantization levels. %
The distributions are poorly matched; this is also seen in ~\cite{Miyashita2016}.
The NUC captures this, since
the complexity for the octave/octave unit ($\max(32, 8) + 4 * 32 + 15 +
3 - 2 = 176$) is less than that of the
octave/linear unit ($32 * 8 + 15 - 1 = 270$).  (And the same for the NWNC.)
If we think of LUT size as an indicator for neural complexity,
we notice that because of the constraints,
the effective size of the LUT is actually much smaller.
Therefore, the representational complexity of the unit is effectively
smaller. %
In summary, we can replace all multiplications
with additions and further reduce our minimal memory requirements but,
when compared on the basis of quantization levels alone, there is
a performance degradation of 5.6\% or more.

Finally, comparing our AlexNet results to those in~\cite{Miyashita2016}:
with $N_{q;w} = 2$ and $N_{q;a} = 1$, we get 79.0\% recall@5
while~\cite{Miyashita2016} gets 75.1\% recall@5 when using
octave/octave quantization in that operating regime.  Additionally,
Figure~\ref{fig:AlexNet Results} shows that the peak performance of
this octave/octave neural unit was better than %
the octave/linear neural unit {\em at the same NUC}
but worse than the octave/linear neural unit 
when compared at the same total number of activation levels.

\section{Large-Scale Experiments}
\label{sec:results}

In this section, we further test our model-free and
octave-based approaches. %
We use AlexNet,
ResNet-101, and MobileNet. We create a ``pre-trained''
model of each network from scratch with ReLU and with ReLU6 units.  All models are trained
on the ImageNet 2012 training set and recall@1 and @5 are computed on
the validation set using the 10-crop procedure~\cite{krizhevsky2012imagenet}. For these experiments, the quantization networks are
fine-tuned from the pre-trained model. The baselines are the  ReLU6
models.

{\em AlexNet}: AlexNet is most useful for comparison to prior work.
  We %
  follow the training procedures specified in~\cite{krizhevsky2012imagenet},
  except: we use RMSProp; initial weight sd=0.005; bias initializer sd=0.1; one Tesla-V100
  GPU; and a stepwise decaying learning rate (LR).  Our ReLU network achieved accuracies:
  recall@1 of 57.4\% and recall@5 of 80.4\%.
  ~\cite{krizhevsky2012imagenet} reported  59.3\% and 81.8\%, recall@1 and
  @5, respectively.  The small difference %
  was because we did not use the PCA pre-processing or LRN~\cite{krizhevsky2012imagenet}.
  Our ReLU6 baseline results: 56.4\% recall@1 and 79.8\% recall@5.
  Fine-tuning is done from this model with LR starting at 0.0001.
  
{\em ResNet-101}: We selected ResNet~\cite{he2015resnet}
  as a modern deep network that takes full advantage of non-separable
  convolutional kernels. For our implementation:  %
  8 GPUs splitting a mini-batch size of 256; and cosine rate
  decay~\cite{Loschilov2016} ending after 300,000 steps.
  Our ReLU model achieved a recall@1 accuracy of 77.8\% and recall@5
  accuracy of 94.1\%. %
  ~\cite{he2015resnet} reported accuracies of 78.25\% and 93.95\%.
  Training with ReLU6, our baseline, we see 74.8\% and 92.3\%.
  We fine-tune from the ReLU network with LR starting at 0.0001.
  
{\em MobileNet}: MobileNet is a compact network for mobile deployment.
  We followed ~\cite{Howard2017} except:
  4 GPUs splitting a mini-batch size of 96;
  and a LR-decay of 0.98 every 13346 steps.
  Our network had a recall@1 accuracy of 68.4\% and @5
  of 88.0\%. ~\cite{Howard2017} reported 70.6\% recall@1.
  Our baseline
  ReLU6 net saw 68.5\% @1 and 88.0\% @5. Fine-tuning
  is from this network %
  with LR-cosine decay starting at 0.0001 for 500,000 steps.

Figure~\ref{fig:AlexNet Results} compares the accuracy of our different
approaches (model-free, octave/linear, and octave/octave) based on the
NUC degrees of freedom in each layer. %
The plot shows the ``efficient
frontier'' for the trade-off in the way that the degrees of freedom
are allocated.  For example, in the AlexNet results, for the
model-free approach, when the NUC is below 1024 entries, it is best to
use 8 activation levels, above that, it uses 16 levels (through 4096
entries), then 32 levels.  Other allocations of NUC did not do as
well.
Even though
the octave/octave has the fewest degrees of freedom, it achieves the
highest recall@1 (and recall@5) accuracy on both AlexNet and ResNet.
Octave/linear gives slightly %
better recall@1 accuracy on
MobileNet but does so using $3\times$ the degrees of freedom.
Similar changes in the best activation quantization levels
were seen in octave/linear and octave/octave results.

For octave/linear results, in the range of 4--32 weights/octave, we
found that the most effective $N_{o;w}$ was closely tied to the step
size used for the activation quantization. 
If a non-zero weight was so small 
that having it multiply the maximum activation value would only
move the accumulated result by at most $\frac{1}{2}$ the activation
step size, changing that weight to zero did not make a measurable
degradation to the overall accuracy.
Instead, it was better to reduce 
$N_{o;w}$ for that choice of activation quantization.  We observed
similar trends for the octave/octave results.%

Table~\ref{table: compare} provides a thorough comparison of our
results with prior work in quantizing both activations and weights
on AlexNet, ResNet, or MobileNet.  After the name column, we use eight
columns to describe the quantization paradigm.  In the first column,
we show whether the approach used bit-shifts or floating/fixed-point multiplies.
In column ``Quant first, last, bias'', we
specify exactly what was quantized - were the first and last layer and
bias unit quantized?  Then we give the NUC (and the per-layer
LUT size, if it differs from the NUC) and the NWNC (and the network-wide combined LUT size, if it differs) as indications of the degrees of freedom and
memory constraints.  The next 2 columns give the quantization levels
for the weights and activations.  ``Per-unit $N_w$ / $N_a$'' are
the numbers of distinct weight \& activation levels
{\em seen by any unit in the network}.  ``Full-net $N_w / N_a$'' are
the numbers of distinct weight \& activation levels
{\em across the full network}.

The results are shown in the the last two sections of
the table (last 6 columns).  They give the baseline performance of all
the methods and how their quantization affects 
performance - measured at recall @1 and @5.  Though
there is an overwhelming amount of information in the table, perhaps
most interestingly, our quantized results often either remain consistent or even
\emph{improve} over the ReLU6 baseline. %

Note that the detailed characterization of the the quantization of
different approaches is necessary due to the breadth of approaches to
quantization.  For all our networks except model-free, the
distinct quantization levels are shared
{\em throughout the whole network}.
(Model-free quantizes each layer independently.)  Many of the previous
approaches to quantizing AlexNet %
use only two or three levels
per layer but then scale each of the eight layers
separately~\cite{Zhou2016,wu2018training,rastegari2016, Hubara2016}. %

To summarize,  all three of our approaches were better than 
previous results on AlexNet.  For ResNet and MobileNet, our results
are nearly identical to those of~\cite{Jacob2018}: we are 0.1\% better
on MobileNet but 0.1\% worse on ResNet.  We use fewer levels (both
weights and activations) than were used in~\cite{Jacob2018} but, using
the freedom given by our table-based approach
(Figure~\ref{fig:methodNoMult}), we are able to place our levels more
strategically than is possible using simple fixed-point
representations.  Remember, also, that our table-based approach 
removes all multiplications,
unlike~\cite{Jacob2018}, opening up highly efficient
FPGA implementations~\cite{Garland2017}.

Finally, two other interesting results from  MobileNet are the
extremely compact versions that are listed on lines 22-23 of
Table~\ref{table: compare}.  Using absolutely no multiplication or
floating point, using only 40 entries (across the two LUTs), we
achieved 66.9\% recall@1: just 1.6\% below our baseline.  If we allow
ourselves 64 entries, we cut the loss in half for 67.7\% recall@1,
0.8\% below our baseline.  With these versions, we are well within
what can fit into on-chip registers.

\section{Conclusions} %
\label{sec:conclude}

Our studies have focused on quantifying neural-unit
complexity (NUC) and network-wide non-compactness (NWNC).  We examined
how to best lower both of these measures without 
impacting classification accuracy.  Our definitions of NUC \& NWNC are very
closely tied to the number of trainable parameters and unique values
{\em at each layer} and {\em across the
  full network}, respectively.  We obtained  improved results over the numerous
previous approaches we examined, even though we quantized more of the
network -- we \emph{completely} quantized the weights, the biases and
the quantizations.  Perhaps what is the most surprising aspect of this
study is the small number
of unique values needed to represent the weights and activations
across the entire network.  
For AlexNet, we achieve 56.5\%
recall@1 and 80.0\% recall@5 with only 347 unique values.  On
ResNet, our best performance is 76.5\% recall@1 and 93.3\% recall@5,
again with only 347 values.  On MobileNet, our two best
results are 69.3\% recall@1 / 88.5\% recall@5 with 338 values
and 69.4\% / 88.5\% with 1038 values.  Using
our table-based approach, %
these
resulting tiny tables allow for in-memory caching %
and require no floating point representation.   Additionally, as
described, we can eliminate all multiplications as well.   These are
all achieved 
without sacrificing classification performance.

\clearpage
\begin{sidewaystable}
\vspace*{3.4in}

\caption{Accuracy under Quantization: Comparison to Prior Work.}
\begin{small}
Key for the ``multiplies?'' column: {\em FP} = some floating-point multiplies in inference;
  {\em fp} = fixed-point multiplies; {\em $\ll$} = scaling by a power of two (\emph{e.g.} bit shifts).
  Key for ``NUC [/$||\text{LUT}||_{lay}$]'' column: neural-unit capacity, in terms of
  trainable parameters, accessible at a single layer
   and the per-layer LUT size, if it differs.
  Key for ``NWNC [/ $||\text{LUT}||_{net}$] column: network-wide non-compactness and network-wide
  count of distinct LUT entries, if it differs; {\em +FP} = some floating-point
  portions not accounted for in the trainable-parameters count. (See
  main text for definitions.)

\label{table: compare}
\begin{center}
\begin{tabular}{r|l|c|p{0.35in}|c|c|c|c||c|c|c||c|c|c|}\cline{2-14}
 & & \multicolumn{6}{c||}{Quantization Description} & \multicolumn{3}{c||}{Recall@1} & \multicolumn{3}{c|}{Recall@5} \\\cline{3-14}
 & & \multicolumn{1}{m{0.35in}|}{mult\-iplies?}
   & \multicolumn{1}{m{0.35in}|}{Quant first,last, (bias$<2^{32}$)}
   & \multicolumn{1}{m{0.5in}|}{NUC {\footnotesize \hspace*{-0.1in} [/$||\text{LUT}||_{lay}$ if differ]}}
   & \multicolumn{1}{m{0.5in}|}{NWNC {\footnotesize \hspace*{-0.1in} [/$||\text{LUT}||_{net}$ if differ]}} &
   \multicolumn{1}{m{0.5in}|}{per-unit $N_w$ / $N_a$} &
   \multicolumn{1}{m{0.5in}|}{full-net $N_w$ / $N_a$} &
   \multicolumn{1}{m{0.25in}|}{base\-line} & quant. & diff. &
   \multicolumn{1}{m{0.25in}|}{base\-line} & quant. & diff. \\\cline{2-14}
 \multicolumn{14}{c}{ } \\\hline\hline
 \multicolumn{14}{c}{%
 \normalsize AlexNet} \\\cline{2-14}
1 & \underline{Our work} & & & & & & & 56.4\% & & & 79.8\% & & \\
 2 & ~ ~ Model-free & {\em none} & \checkmark \checkmark \checkmark &
  16,384 & 131,072 & 512 / 32 & 4096 / 32 &
   & 56.5\% & +0.1\% & & 79.9\% & +0.1\%
  \\
 3 & ~ ~ Octave/linear & $\ll$ & \checkmark \checkmark \checkmark &
   1038 / 1024 & 1038 / 1024 & 481 / 64 & 481 / 64 &
   & 56.5\% & +0.1\% & & 79.8\% & 0.0\%
   \\
 4 & ~ ~ Octave/octave & $\ll$ & \checkmark \checkmark \checkmark &
   347 / 320 & 347 / 320 & 385 / 321 & 385 / 321 &
   & {\bf 56.6\%} & {\bf +0.2\%} & & {\bf 80.0\%} & {\bf +0.2\%}
   \\\cline{2-14}

 5 &  \cite{wu2018training} & FP & \checkmark \xmark \checkmark &
   512 or FP & 526 + FP & 3 / 255 & 24 / 2040 &
    - & - & - & {\bf 80.7\%} & 75.9\% & -4.8\% \\\cline{2-14}
 6 &   \cite{Miyashita2016}$^\dagger$ & $\ll$ &
    \checkmark \checkmark \checkmark  & {\bf 288} / {\bf 4} &
    {\bf 296} / {\bf 32} & 256 / 16 & 1,632  / 128  &
      - & - & - & 78.3\% & 75.1\% & -2.7\% \\\cline{2-14}
 7 &   \cite{Zhou2016}$^\ddagger$ & FP & \xmark \xmark \xmark
 & 65,536 + FP & 65,550 + FP &
 256 / 256 & 2048 / 2048 &
      55.9\% & 53.0\% & -2.9\% & - & - & - \\\cline{2-14}
 8 &   \cite{Hubara2016}$^\sharp$ & FP & \checkmark \checkmark \xmark &
    8 or FP & 22 + FP & 2 / 4 & 16 / 32 &
      {\bf 56.6\%} & 51.0\% & -5.6\% & 80.2\% & 73.7\% & -6.5\% \\\cline{2-14}
 9 &   \cite{rastegari2016} & FP & \xmark \xmark  \checkmark &
    4 or FP & 18 + FP & 2 / 2 & 16 / 16 &
      {\bf 56.6\%} & 44.2\% & -12.4\% & 80.2\% & 69.2\% & -11.0\%\\\cline{2-14}
 \multicolumn{14}{c}{ } \\\hline\hline
 \multicolumn{14}{c}{%
 \normalsize ResNet} \\\cline{2-14}
10 & \underline{Our work} & & & & & & &
 74.8\% & & & {\bf 92.3\%} & & \\
11 & ~ ~ Model-free/linear & {\em none} & \checkmark \checkmark \checkmark &
   16,384 & 1,654,784 & 512 / 32 & 51,712 / 32 &
     & 76.3\% & +1.5\% & & 93.2\% & +0.7\% \\
12 & & {\em none} & \checkmark \checkmark \checkmark &
   8,192 & 827,392 & 256 / 32 & 25,856 / 32 &
     & 75.9\% & +1.1\% & & 93.0\% & +0.7\% \\
13 & ~ ~ Octave/linear & $\ll$ & \checkmark \checkmark \checkmark &
   2062 / 2048 & 2062 / 2048 & 961 / 64 & 961 / 64 &
   & 76.2\% & +1.4\% & & 93.2\% & +0.9\%  \\
14 &  & $\ll$ & \checkmark \checkmark  \checkmark &
  526 / 512 & 526 / 512 & 241 / 64 & 241 / 64 &
    & 75.2\% & +0.4\% & & 92.7\% & +0.4\% \\
15 & ~ ~ Octave/octave & $\ll$ & \checkmark \checkmark \checkmark &
  {\bf 347} / {\bf 320} &  {\bf 347} / {\bf 320} & 1537 / 321 & 1537 / 321 &
  & 76.5\% & {\bf +1.7\%} & & {\bf 93.3\%} & {\bf +1.0\%} \\\cline{2-14}
16 &  ~\cite{Jacob2018} & fp & \checkmark \checkmark \xmark &
  65,536 & 65,736 & 256 / 256 & \multicolumn{1}{m{0.5in}||}{25,856 / 25,856} &
      {\bf 78.0\%} & {\bf 76.6\%} & -1.4\% & - & - & - \\\cline{2-14}
  \multicolumn{14}{c}{ } \\\hline\hline
  \multicolumn{14}{c}{%
  \normalsize MobileNet} \\\cline{2-14}
 17 & \underline{Our work} & & & & & & &
   68.5\% & & & {\bf 88.0\%} & & \\
18 & ~ ~ Model-free/linear & {\em none} & \checkmark \checkmark \checkmark &
  16,384 & 1,835,008 & 256 / 64 & 7,168 / 64 &
  & 69.3\% & +0.8\% & & {\bf 88.6\%} & {\bf +0.6\%} \\
19 & ~ ~ Octave/linear & $\ll$ & \checkmark \checkmark \checkmark &
  1038 / 1024 & 1038 / 1024 & 482 / 64 & 481 / 64 &
  & {\bf 69.4\%} & {\bf +0.9\%} & & 88.5\% & +0.5\%  \\
20 &    & $\ll$ & \checkmark \checkmark  \checkmark &
  526 / 512 & 526 / 512 & 241 / 64 & 241 / 64 &
    & 69.1\% & +0.6\% & & 88.4\% & +0.4\%  \\
21 & ~ ~ Octave/octave & $\ll$ & \checkmark \checkmark \checkmark &
   338 / 320 & 338 / 320 & 961 / 321 & 961 / 321 &
    & 69.3\% & +0.8\% & & 88.5\% & +0.5\% \\
22 &    & $\ll$ & \checkmark \checkmark  \checkmark &
   {\bf 73} / {\bf 40} & {\bf 73} / {\bf 40} & 497 / 33 & 497 / 33 &
    & 66.9\% & -1.6\% & & 86.9\% & -1.1\% \\
23 &   & $\ll$ & \checkmark \checkmark  \checkmark &
   97 / 64 & 97 / 64 & 1985 / 33 & 1985 / 33 &
    & 67.7\% & -0.8\% & & 87.3\% & -0.7\% \\
    \cline{2-14}
24 & ~\cite{Jacob2018}$^{\natural}$ & fp & \checkmark \checkmark  \xmark &
    65,536 & 65,592 & 256 / 256 & 7,168 / 7,168 &
    {\bf 70.5\%} & 69.3\% & -1.2\% & - & - & - \\\cline{2-14}
    
     \multicolumn{14}{c}{ } \\
     \multicolumn{14}{l}{\footnotesize ~ ~ ~ $^\dagger$ For~\cite{Miyashita2016}, NUC is for Conv3 layer, the widest and full-net $N_a$ and $N_w$ include both numbers of octaves and per-layer scaling.} \\
     \multicolumn{14}{l}{\footnotesize ~ ~ ~ $^\ddagger$ For~\cite{Zhou2016}: Bias-count answer based on \cite{alexnet-dorefa}.} \\
     \multicolumn{14}{l}{\footnotesize ~ ~ ~ $^\sharp$ For~\cite{Hubara2016}:
    FP operations needed in the ``approximate batch-norm'' (AP2) that
    they add before the first layer.} \\
\end{tabular}
\end{center}
\end{small}

\end{sidewaystable}

\clearpage

\bibliography{binary}

\begin{thebibliography}{49}
\providecommand{\natexlab}[1]{#1}
\providecommand{\url}[1]{\texttt{#1}}
\expandafter\ifx\csname urlstyle\endcsname\relax
  \providecommand{\doi}[1]{doi: #1}\else
  \providecommand{\doi}{doi: \begingroup \urlstyle{rm}\Url}\fi

\bibitem[Achterhold et~al.(2018)Achterhold, Koehler, Schmeink, and
  Genewein]{Achterhold2018}
Achterhold, J., Koehler, J.~M., Schmeink, A., and Genewein, T.
\newblock Variational network quantization.
\newblock In \emph{International Conference on Learning Representations
  (ICLR)}, 2018.

\bibitem[Anwar et~al.(2015)Anwar, Hwang, and Sung]{Anwar2015}
Anwar, S., Hwang, K., and Sung, W.
\newblock Fixed point optimization of deep convolutional neural networks for
  object recognition.
\newblock In \emph{Proceedings of the IEEE International Conference on
  Acoustics, Speech and Signal Processing (ICASSP)}, 2015.

\bibitem[Baluja et~al.(2018)Baluja, Marwood, Covell, and Johnston]{Baluja2018b}
Baluja, S., Marwood, D., Covell, M., and Johnston, N.
\newblock No multiplication? no floating point? no problem! training networks
  for efficient inference.
\newblock \emph{arXiv preprint}, 2018.

\bibitem[Balzer et~al.(1991)Balzer, Takahashi, Ohta, and Kyuma]{Balzer1991}
Balzer, W., Takahashi, M., Ohta, J., and Kyuma, K.
\newblock Weight quantization in {B}oltzmann machines.
\newblock \emph{Neural Networks}, 4\penalty0 (3):\penalty0 405--409, 1991.

\bibitem[Cai et~al.(2017)Cai, He, Sun, and Vasconcelos]{Cai2017}
Cai, Z., He, X., Sun, J., and Vasconcelos, N.
\newblock Deep learning with low precision by half-wave gaussian quantization.
\newblock In \emph{Computer Vision and Pattern Recognition (CVPR)}, 2017.

\bibitem[Chatterjee(2018)]{Chatterjee2018}
Chatterjee, S.
\newblock Learning and memorization.
\newblock In \emph{International Conference on Machine Learning}, 2018.

\bibitem[Choi et~al.(2017)Choi, El-Khamy, and Lee]{Choi2017}
Choi, Y., El-Khamy, M., and Lee, J.
\newblock Towards the limit of network quantization.
\newblock In \emph{International Conference on Learning Representations
  (ICLR)}, 2017.

\bibitem[Courbariaux et~al.(2015)Courbariaux, Bengio, and
  David]{Courbariaux2015}
Courbariaux, M., Bengio, Y., and David, J.-P.
\newblock {BinaryConnect}: Training deep neural networks with binary weights
  during propagations.
\newblock In \emph{Advances in Neural Information Processing Systems}, 2015.

\bibitem[Courbariaux et~al.(2016)Courbariaux, Hubara, Soudry, El-Yaniv, and
  Bengio]{courbariaux2016}
Courbariaux, M., Hubara, I., Soudry, D., El-Yaniv, R., and Bengio, Y.
\newblock Binarized neural networks: Training deep neural networks with weights
  and activations constrained to +1 or-1.
\newblock \emph{arXiv preprint arXiv:1602.02830}, 2016.

\bibitem[Deng et~al.(2009)Deng, Dong, Socher, Li, Li, and
  Fei-Fei]{deng2009imagenet}
Deng, J., Dong, W., Socher, R., Li, L.-J., Li, K., and Fei-Fei, L.
\newblock Imagenet: A large-scale hierarchical image database.
\newblock In \emph{Computer Vision and Pattern Recognition, 2009. CVPR 2009.
  IEEE Conference on}, pp.\  248--255. IEEE, 2009.

\bibitem[Deng et~al.(2017)Deng, Jiao, Pei, Wu, and Li]{deng2017}
Deng, L., Jiao, P., Pei, J., Wu, Z., and Li, G.
\newblock Gated {XNOR} networks: Deep neural networks with ternary weights and
  activations under a unified discretization framework.
\newblock \emph{CoRR}, abs/1705.09283, 2017.
\newblock URL \url{http://arxiv.org/abs/1705.09283}.

\bibitem[Garland \& Gregg(2017)Garland and Gregg]{Garland2017}
Garland, J. and Gregg, D.
\newblock Low complexity multiply-accumulate units for convolutional neural
  networks with weight-sharing.
\newblock \emph{IEEE Computer Architecture Letters}, 2017.

\bibitem[Gong et~al.(2014)Gong, Liu, Yang, and Bourdev]{Gong2014}
Gong, Y., Liu, L., Yang, M., and Bourdev, L.
\newblock Compressing deep convolutional networks using vector quantization.
\newblock \emph{arXiv 1412.6115}, 2014.

\bibitem[Guo(2018)]{Guo2018}
Guo, Y.
\newblock A survey on methods and theories of quantized neural networks.
\newblock \emph{arXiv 1808.04752}, 2018.

\bibitem[Han et~al.(2015)Han, Mao, and Dally]{Han2015}
Han, S., Mao, H., and Dally, W.
\newblock Deep compression: Compressing deep neural networks with pruning,
  trained quantization and huffman coding.
\newblock \emph{arXiv 1510.00149}, 2015.

\bibitem[Han et~al.(2016)Han, Mao, and Dally]{nvidia2016}
Han, S., Mao, H., and Dally, W.~J.
\newblock Deep compression: Compressing deep neural networks with pruning,
  trained quantization and huffman coding.
\newblock \emph{International Conference on Learning Representations (ICLR)},
  2016.

\bibitem[He et~al.(2015)He, Zhang, Ren, and Sun]{he2015resnet}
He, K., Zhang, X., Ren, S., and Sun, J.
\newblock Deep residual learning for image recognition.
\newblock \emph{CoRR}, abs/1512.03385, 2015.
\newblock URL \url{http://arxiv.org/abs/1512.03385}.

\bibitem[Hinton(2012)]{hinton2012lecture}
Hinton, G.
\newblock Neural networks for machine learning.
\newblock Coursera, video lectures, 2012.

\bibitem[Howard et~al.(2017)Howard, Zhu, Chen, Kalenichenko, Wang, Weyand,
  Andreetto, and Adam]{Howard2017}
Howard, A., Zhu, M., Chen, B., Kalenichenko, D., Wang, W., Weyand, T.,
  Andreetto, M., and Adam, H.
\newblock Mobilenets: Efficient convolutional neural networks for mobile vision
  applications.
\newblock \emph{arXiv 1704.04861}, 2017.

\bibitem[Hubara et~al.(2016)Hubara, Courbariaux, Soudry, El-Yaniv, and
  Bengio]{Hubara2016}
Hubara, I., Courbariaux, M., Soudry, D., El-Yaniv, R., and Bengio, Y.
\newblock Quantized neural networks: Training neural networks with low
  precision weights and activations.
\newblock \emph{arXiv arXiv:1609.07061}, 2016.

\bibitem[Hwang \& Sung(2014)Hwang and Sung]{Hwang2014}
Hwang, K. and Sung, W.
\newblock Fixed-point feedforward deep neural network design using weights
  {+1}, 0, and {-1}.
\newblock In \emph{Workshop on Signal Processing Systems (SiPS)}, 2014.

\bibitem[Ioffe \& Szegedy(2015)Ioffe and Szegedy]{Ioffe2015}
Ioffe, S. and Szegedy, C.
\newblock Batch normalization: Accelerating deep network training by reducing
  internal covariate shift.
\newblock In \emph{Proc. International Conference on Machine Learning}, pp.\
  448--456, 2015.

\bibitem[Jacob et~al.(2018)Jacob, Kligys, Chen, Zhu, Tang, Howard, Adam, and
  Kalenichenko]{Jacob2018}
Jacob, B., Kligys, S., Chen, B., Zhu, M., Tang, M., Howard, A., Adam, H., and
  Kalenichenko, D.
\newblock Quantization and training of neural networks for efficient
  integer-arithmetic-only inference.
\newblock In \emph{Computer Vision and Pattern Recognition (CVPR)}, 2018.

\bibitem[Jain(2010)]{jain2010data}
Jain, A.~K.
\newblock Data clustering: 50 years beyond k-means.
\newblock \emph{Pattern recognition letters}, 31\penalty0 (8):\penalty0
  651--666, 2010.

\bibitem[Kim et~al.(2014)Kim, Hwang, and Sung]{Kim2014}
Kim, J., Hwang, K., and Sung, W.
\newblock X1000 real-time phoneme recognition {VLSI} using feed-forward deep
  neural networks.
\newblock In \emph{International Conference on Acoustics, Speech and Signal
  Processing (ICASSP)}, 2014.

\bibitem[Krishnamoorthi(2018)]{Krishnamoorthi2018}
Krishnamoorthi, R.
\newblock Quantizing deep convolutional networks for efficient inference: A
  whitepaper.
\newblock \emph{arXiv 1806.08342}, 2018.

\bibitem[Krizhevsky et~al.(2012)Krizhevsky, Sutskever, and
  Hinton]{krizhevsky2012imagenet}
Krizhevsky, A., Sutskever, I., and Hinton, G.~E.
\newblock Imagenet classification with deep convolutional neural networks.
\newblock In \emph{Advances in neural information processing systems}, pp.\
  1097--1105, 2012.

\bibitem[Langhammer \& Baeckler(2018)Langhammer and Baeckler]{Langhammer2018}
Langhammer, M. and Baeckler, G.
\newblock High density and performance multiplication for fpga.
\newblock In \emph{IEEE Symposium on Computer Arithmetic}, 2018.

\bibitem[Li et~al.(2017)Li, De, Xu, Studer, Samet, and Goldstein]{Li2017}
Li, H., De, S., Xu, Z., Studer, C., Samet, H., and Goldstein, T.
\newblock Training quantized nets: A deeper understanding.
\newblock In \emph{Advances in Neural Information Processing Systems}, 2017.

\bibitem[Lin et~al.(2015)Lin, Talathi, and Annapureddy]{qualcomm2017}
Lin, D.~D., Talathi, S.~S., and Annapureddy, V.~S.
\newblock Fixed point quantization of deep convolutional networks.
\newblock \emph{CoRR}, abs/1511.06393, 2015.
\newblock URL \url{http://arxiv.org/abs/1511.06393}.

\bibitem[Loschilov \& Hutter(2016)Loschilov and Hutter]{Loschilov2016}
Loschilov, I. and Hutter, F.
\newblock Sgdr: Stochastic gradient descent with warm restarts.
\newblock \emph{arXiv 1608.03983}, 2016.

\bibitem[Marchesi et~al.(1993)Marchesi, Orlandi, Piazza, and
  Uncini]{Marchesi1993}
Marchesi, M., Orlandi, G., Piazza, F., and Uncini, A.
\newblock Fast neural networks without multipliers.
\newblock \emph{IEEE transactions on Neural Networks}, 4\penalty0 (1):\penalty0
  53--62, 1993.

\bibitem[Marcus(2004)]{Marcus2004}
Marcus, G.
\newblock Open cores project fpuvhdl: Floating point adder and mulitplier,
  2004.
\newblock URL \url{https://opencores.org/projects/fpuvhdl}.

\bibitem[Miyashita et~al.(2016)Miyashita, Lee, and Murmann]{Miyashita2016}
Miyashita, D., Lee, E., and Murmann, B.
\newblock Convolutional neural networks using logarithmic data representation.
\newblock \emph{arXiv 1603.01025}, 2016.

\bibitem[Rastegari et~al.(2016)Rastegari, Ordonez, Redmon, and
  Farhadi]{rastegari2016}
Rastegari, M., Ordonez, V., Redmon, J., and Farhadi, A.
\newblock {XNOR}-net: Imagenet classification using binary convolutional neural
  networks.
\newblock In \emph{European Conference on Computer Vision}, pp.\  525--542.
  Springer, October 2016.

\bibitem[Sabeetha et~al.(2015)Sabeetha, Ajayan, Shriram, Vivek, and
  Rajesh]{Sabeetha2015}
Sabeetha, S., Ajayan, J., Shriram, S., Vivek, K., and Rajesh, V.
\newblock A study of performance comparison of digital multipliers using 22nm
  strained silicon technology.
\newblock In \emph{Proc. International Conference on Electronics and
  Communication Systems (ICECS)}, pp.\  180–184, 2015.

\bibitem[Salimans \& Kingma(2016)Salimans and Kingma]{Salimans2016}
Salimans, T. and Kingma, D.~P.
\newblock Weight normalization: A simple reparameterization to accelerate
  training of deep neural networks.
\newblock In \emph{Proceedings of the 30th International Conference on Neural
  Information Processing Systems}, NIPS'16, 2016.
\newblock URL \url{http://dl.acm.org/citation.cfm?id=3157096.3157197}.

\bibitem[Tang \& Kwan(1993)Tang and Kwan]{Tang1993}
Tang, C.~Z. and Kwan, H.~K.
\newblock Multilayer feedforward neural networks with single powers-of-two
  weights.
\newblock \emph{IEEE Transactions on Signal Processing}, 41\penalty0
  (8):\penalty0 2724--2727, 1993.

\bibitem[Tang et~al.(2017)Tang, Hua, and Wang]{Tang2017}
Tang, W., Hua, G., and Wang, L.
\newblock How to train a compact binary neural network with high accuracy?
\newblock In \emph{AAAI Conference on Artificial Intelligence}, 2017.

\bibitem[Vanhoucke et~al.(2011)Vanhoucke, Senior, and Mao]{Vanhoucke2011}
Vanhoucke, V., Senior, A., and Mao, M.
\newblock Improving the speed of neural networks on cpus.
\newblock In \emph{NIPS Workshop on Deep Learning and Unsupervised Feature
  Learning}, 2011.

\bibitem[Wang \& Song(2011)Wang and Song]{wang2011ckmeans}
Wang, H. and Song, M.
\newblock Ckmeans. 1d. dp: optimal k-means clustering in one dimension by
  dynamic programming.
\newblock \emph{The R journal}, 3\penalty0 (2):\penalty0 29, 2011.

\bibitem[Warren(2013)]{Warren2013}
Warren, Jr., H.
\newblock Hacker's delight: Number of leading zeros, 2013.
\newblock URL \url{http://www.hackersdelight.org/hdcodetxt/nlz.c.txt}.

\bibitem[Wu et~al.(2018)Wu, Li, Chen, and Shi]{wu2018training}
Wu, S., Li, G., Chen, F., and Shi, L.
\newblock Training and inference with integers in deep neural networks.
\newblock \emph{arXiv preprint arXiv:1802.04680}, 2018.

\bibitem[Wu \& Zou(2018)Wu and Zou]{alexnet-dorefa}
Wu, Y. and Zou, Y.
\newblock tensorpack/examples/dorefa-net/alexnet-dorefa.py - github.
\newblock
  https://github.com/tensorpack/tensorpack/blob/master/examples/DoReFa-Net/alexnet-dorefa.py,
  2018.

\bibitem[Xilinx(2018)]{Xilinx2018}
Xilinx.
\newblock 7 series fpgas data sheet: Overview (ds180), 2018.
\newblock URL
  \url{https://www.xilinx.com/support/documentation/data_sheets/ds180_7Series_Overview.pdf}.

\bibitem[Yi et~al.(2008)Yi, Hangping, and Bin]{yi2008new}
Yi, Y., Hangping, Z., and Bin, Z.
\newblock A new learning algorithm for neural networks with integer weights and
  quantized non-linear activation functions.
\newblock In \emph{IFIP International Conference on Artificial Intelligence in
  Theory and Practice}, pp.\  427--431. Springer, 2008.

\bibitem[Zhou et~al.(2017)Zhou, Yao, Guo, Xu, and Chen]{Zhou2017}
Zhou, A., Yao, A., Guo, Y., Xu, L., and Chen, Y.
\newblock Incremenal network quantization: Towards lossless cnns with
  low-precision weights.
\newblock In \emph{International Conference on Learning Representations
  (ICLR)}, 2017.

\bibitem[Zhou et~al.(2016)Zhou, Wu, Ni, Zhou, Wen, and Zou]{Zhou2016}
Zhou, S., Wu, Y., Ni, Z., Zhou, X., Wen, H., and Zou, Y.
\newblock {DoReFa}-net: Training low bitwidth convolutional neural networks
  with low bitwidth gradients.
\newblock \emph{arXiv arXiv:1606.06160}, June 2016.

\bibitem[Zhu et~al.(2016)Zhu, Han, Mao, and Dally]{Zhu2016}
Zhu, C., Han, S., Mao, H., and Dally, W.~J.
\newblock Trained ternary quantization.
\newblock \emph{arXiv 1612.01064}, 2016.

\end{thebibliography}
\bibliographystyle{icml2018}

\clearpage

\appendix

\section{Quantization with Batch-Norm and Weight Norm: Implementation Details}
\label{supplemental sec:norm}

As pointed out in Section~\ref{sec:norm}, 
Batch norm (BN)~\cite{Ioffe2015} makes it difficult to train quantized
networks since normalization moves the
weights and biases of each unit independently. To avoid the
compromises that an ongoing BN process would impose on our quantized
network, we instead train with BN in the continuous-weight domain and
only work towards a quantized network after the continuous-weight
version has converged. Once the network has converged, we fold the
changes dictated by the normalization operations into the
weight-layers immediately preceding or following the normalization
functions.

Using a variation on the notation from~\cite{Ioffe2015},
BN changes each input $x$ to $y$ via
\begin{eqnarray*}
  y & = & \frac{\gamma}{\sigma} x + (\beta -\frac{\gamma}{\sigma} m) \\
  \sigma & = & \sqrt{\text{Var}[x] + \epsilon} \\
  m & = & E[x] \\
  \gamma, \beta & \text{are} & \text{the BN learned
    parameters}
\end{eqnarray*}
When the BN operation is immediately {\em after} a weight layer,
then that weight layer’s biases ($b$) and weights ($w$) change to
  \begin{eqnarray*}
    b & \leftarrow & \frac{\gamma}{\sigma} b + \beta -
    \frac{\gamma}{\sigma} m \\
    w & \leftarrow & \frac{\gamma}{\sigma} w
\end{eqnarray*}
    
If the BN is immediately {\em before} a weight layer, then there are
likely to be many different BN layers, each feeding into a different
weight connection ($w^+_i$).  Using $i$ on the BN parameters to
distinguish these different normalizations, folding all of the
preceding BN's gives us:
  \begin{eqnarray*}
    b^+ & \leftarrow & b^+ + \Sigma_{i} (\beta_i -
    \frac{\gamma_i}{\sigma_i} m_i) w^+_i \\
    w^+_i & \leftarrow & \frac{\gamma_i}{\sigma_i} w^+_i
  \end{eqnarray*}
  (in that order).

Note that, if the batch norm
feeds a skip connection or a tower split, the weights/biases for
all of the convolutional layers that are directly connected to it need
to be modified in the way just described.

For weight normalization~\cite{Salimans2016}, we can use basically the
same formulas as above, but replacing $\frac{\gamma}{\sigma}$ with the
weight normalization scaling ($s$) and replacing $\beta -
\frac{\gamma}{\sigma} m$ with zero.

At the start of fine-tuning, the model is reconstructed without
batch-norm.  If associated weight layer, in the original network did
not have a bias tensor (which they often do not when associated with a
BN layer), an initially-zero bias tensor is added to the weight
layer. The 
pre-trained batch-norm parameters are folded into the pre-trained
weights and biases. Training continues, with periodic quantization
(according to one of the quantization approaches
described in the main paper) and
without any batch-norm computation.

\section{Weight/bias quantization using a Laplacian distribution model}
\label{supplemental sec:laplacian}

\begin{figure}
\centering

\includegraphics[width=0.3\linewidth]{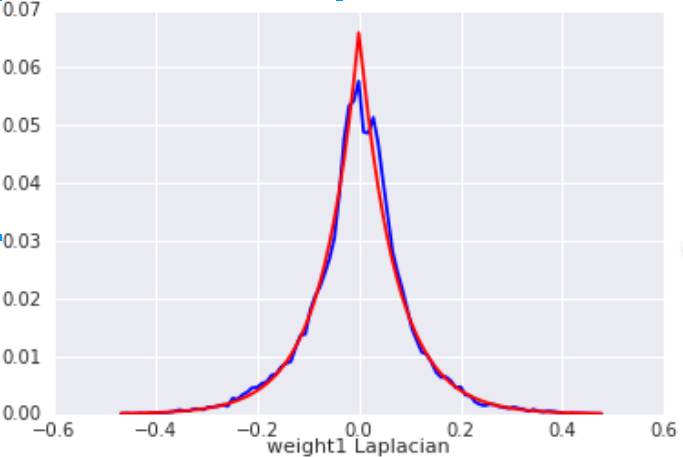}\hfill\includegraphics[width=0.3\linewidth]{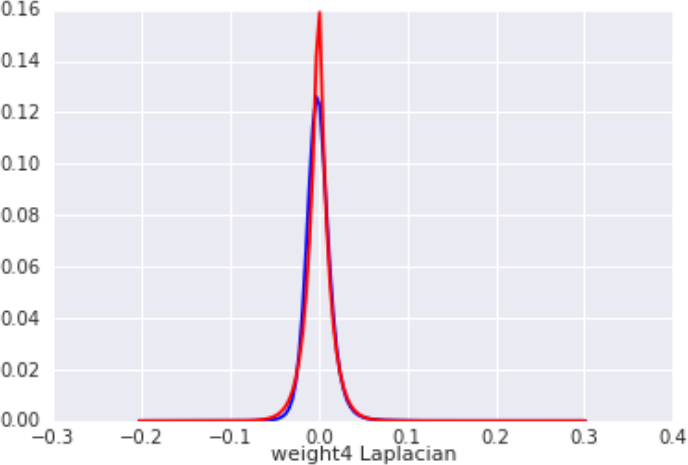}\hfill\includegraphics[width=0.3\linewidth]{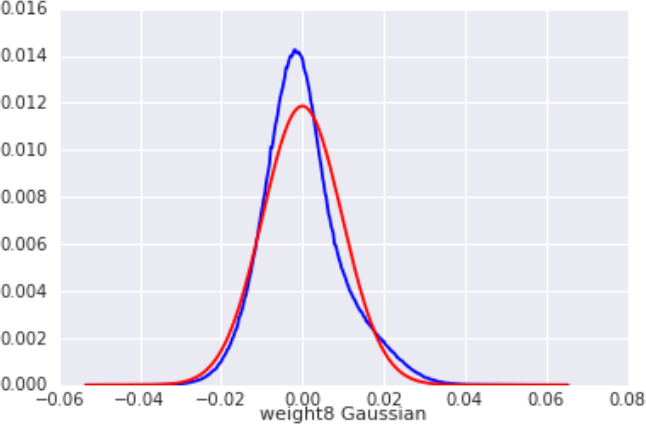}

\caption{Histograms of weights from a well-trained (continuous)
  AlexNet, from layers 1, 4, and 8.  The blue line is the actual weight
  distribution; the red one is the best fitting distribution (either
  Laplacian or Gaussian).  The distributions of layers 2, 3, and 5
  look very similar to the Laplacian distribution of layer 4.  The
  distributions of layers 6  and 7 are similar to the Gaussian
  distribution of layer 8 with smaller variances.
  {\em (Reproduced with permission from~\cite{Baluja2018b})}}
\label{fig:AlexNet histograms}

\end{figure}

As asserted in Section~\ref{sec:laplacian} and shown in
Figure~\ref{fig:AlexNet histograms}, many fully-trained weight
distributions resemble Laplacian or Gaussian distributions.  This
insight can help us to better select cluster centers and cluster
occupancies.

If we have a sample from a Laplacian distribution and if it is
small enough to cluster using the dynamic-programming implementation
of 1-D k-means clustering then that method will provide the optimal
clusters for that sample.  However, with millions of weights (which are
the samples we are trying to cluster), we can not do that.  Instead,
we can find the best clusters for the {\em expected} $L_1$ or $L_2$
error across all fair samples.
Figure~\ref{fig:DistributionModel} (repeated here for ease of
reference) shows 
cluster centers and occupancies for Laplacian distributions that
minimize its $L_1$ or $L_2$ error in the weight/bias space.

One interesting characteristic of the $L_1$ Laplacian-based clustering
model is that we can describe the best cluster center locations in
closed form, as a function of the near-extreme values that were seen
in the weight/bias set.  We can also express the best cluster
occupancies, as a function of the sample size and the number of clusters.
Specifically, for the cluster centers with minimum expected $L_1$ error
(using an odd number, $N$, of cluster centers), the cluster
centers should be at $\pm b L_i$ where
$b$ is a scaling factor and where
\begin{eqnarray*}
  L_i & = & L_{i-1} + \Delta_i \\
  \Delta_i & = & - \ln(1 - 2 \exp(L_{i-1}) / N) \\
  L_0 & = & 0
  \end{eqnarray*}
We set the
scaling factor, $b$, using the cluster occupancy curve for guidance.  From
Figure~\ref{fig:DistributionModel}-b, for minimum $L_1$ error on a
fair sample set from a Laplacian distribution, occupancy of the
clusters should fall linearly from the central peak of $\frac{2}{N}
N_w$ to the outer most bins at $\frac{2}{N^2} N_w$.  Using this
insight, we set
$b = W_{max} / L_{N/2}$ where $W_{max}$ is based on the weight
magnitudes in the outer half of the outermost quantization bins.

We gave the description for the cluster occupancy with the minimum
expected $L_1$ error in Subsection~\ref{sec:model-free}.  Repeating that here:
The cluster occupancy counts which minimize the expected $L_1$
Laplacian error is a 
(discretized) symmetric-about-zero triangle whose width is $N_w + 2$
and whose area is $N_{\text{net}}$.  For the cluster counts (unlike
the cluster centers), there are no scale parameters to be estimated.

Since some of the layers' weight distributions are closer to Gaussian
than Laplacian (for example, layer 8 of AlexNet, shown in
Figure~\ref{fig:AlexNet histograms}), we need to consider this model
mismatch.
However, using the Laplacian
occupancy profile works well for both Laplacian and Gaussian
distributions: the increase in $L_1$ error caused by using the
Laplacian profile on a Gaussian distribution is under
4\%. The
Laplacian occupancy profile does not work as well for heavy-tailed
distributions, like the Cauchy distribution, resulting in 51\% more
error than could have been achieved by using the correct probability
profile.  However, none of the weight distributions that we have
encountered in our network training were heavy tailed.

We give our results using this Laplacian-model approach in
Subsection~\ref{sec:laplacian}.

\section{Pseudo-code for Weight quantization approaches}
\label{supplemental sec:weight quant}

Due to the simplicity of the 1-D k-means weight clustering, we do
not include pseudo-code for that approach.  Since we 
dropped the Laplacian quantization approach for our later studies,
we do not include pseudo-code for that approach, just the mathematical
description in Appendix~\ref{supplemental sec:laplacian}.

\subsection{Weight/bias quantization using a non-parametric probability distribution model}
\label{supplemental sec:model-free-weight-quant}

The code for Section~\ref{sec:model-free} is in Algorithm 1.

\begin{figure*}
  \begin {minipage}{\linewidth}
 \centering
  \begin{algorithm}[H]
    \renewcommand{\algorithmicrequire}{\textbf{Input:}}
    \renewcommand{\algorithmicensure}{\textbf{Output:}}
    \caption{Model-Free Weight Quantization Pseudo-Code: Training}
    Throughout, fully-connected layers are implemented as conv2d
    layers with a 1x1 kernel.
    \begin{algorithmic}[1]
    \item[]
      ~\\
      \COMMENT {****INITIALIZATION****}
    \item[]      
      ~\\
      \REQUIRE $N_q$: The number of weights; $model$: a pre-trained model; $cut\_values$: a dictionary from the conv layer tensors in $model$ to a list of indices.
  \FORALL {$batch\_norm\_layer \in model$}
    \STATE Fold $batch\_norm\_layer$ into the preceding conv layer (see Section 2.3).
  \ENDFOR
  \STATE \COMMENT {$ConvLayerTensors(model)$ returns all the conv layer tensors in model, both weights and biases.}
  \STATE \COMMENT {Build $quantized\_values$, a dictionary from each tensor in $ConvLayerTensors(model)$ to its sorted, quantized values.}
  \STATE \COMMENT {The input $cut\_values[tensor]$ is a list of indices. $cut\_values[tensor][i+1]-cut\_values[tensor][i]$ is the number of values in tensor that belong in bucket $i$, for $i = 0$ to $N_q$.}
  \FORALL {$tensor \in ConvLayerTensors(model)$}
    \STATE $sorted\_tensor \gets sort(tensor.flatten())$
    \FOR {$i = 0$ to $N_q$}
      \STATE $cut\_start = cut\_values[i]$
      \STATE $cut\_end = cut\_values[i]$
      \STATE $centroid \gets average(sorted\_tensor[cut\_start:cut\_end])$
      \STATE $quantized\_values[tensor] \gets quantized\_values[tensor] + [centroid] * (cut\_end - cut\_start)$
    \ENDFOR
  \ENDFOR      
  \ENSURE $quantized\_values$; $model$

    \item[]
      ~\\
      ~\\
      \COMMENT {****EVERY 1000 TRAINING STEPS****}
    \item[]      
      ~\\
  \REQUIRE $quantized\_values$; $model$
  \FORALL {$input\_tensor \in ConvLayerTensors(model)$}
    \STATE \COMMENT {Assign the $quantized\_values$ from the initial model into the tensors.}
    \STATE $arg\_sorted = argsort(input\_tensor.flatten())$
    \STATE $quantized_tensor \gets scatter(quantized\_values[ input\_tensor], arg\_sorted)$
    \STATE $input\_tensor \gets quantized_tensor.reshape(input\_tensor.shape)$
  \ENDFOR
  \STATE Save $model$, the quantized checkpoint.
    
    \end{algorithmic}
  \end{algorithm} 
  \end{minipage}

  \label{fig:model-free-weight-quant} 
   \vspace {0.0in}  
\end{figure*}

\subsection{Weight/bias quantization using Octave}
\label{supplemental sec:octave-weight-quant}

The code for Section~\ref{sec:octave} is in Algorithm 2.

\begin{figure*}[t]
  \begin {minipage}{0.99\linewidth}  
 \centering
  \begin{algorithm}[H]
    \renewcommand{\algorithmicrequire}{\textbf{Input:}}
    \renewcommand{\algorithmicensure}{\textbf{Output:}}
    \caption{Octave Weight Quantization Pseudo-Code}
    Throughout, fully-connected layers are implemented as conv2d
    layers with a 1x1 kernel.
    \begin{algorithmic}[1]
    \item[]
      ~\\
      \COMMENT {**** Compute a codebook of all the quantized values. ****}
    \item[]
      ~\\      
  \REQUIRE $N_o$:  number of octaves;  $N_q$: number of quantization levels within each octave; $K_{max}$:  maximum quantized value

  \STATE $codebook \gets K_{max} \times pow(2, range(-N_q \times N_o - 1, 1)
    / N_q)$
  \STATE $codebook \gets -flip(codebook) + [0] + codebook$\\
  \COMMENT {\textbf{Explanation:} ~Tensor values in the range
    $(cut\_values[i], cut\_values[i+1])$
    will be quantized to $codebook[i+1]$.\\
    ~~~~~~~~~~~~~~~~~~~~~~~~~Values less than $cut\_values[0]$
    are quantized to $codebook[0]$.\\
    ~~~~~~~~~~~~~~~~~~~~~~~~~Values greater than $cut\_values[-1]$
    are quantized to $codebook[-1]$.}
  \STATE $cut\_values \gets [(codebook[i] + codebook[i+1])/2$
    for $i = 0$ to $len(codebook)-1]$
  \ENSURE $codebook, cut\_values$
  \end{algorithmic}
  \end{algorithm}
  \end{minipage}

  \label{fig:octave-weight-quant} 
   \vspace {0.0in}  
\end{figure*}

\section{Addition under octave-based quantization for both
  activation outputs and weights/biases}
\label{supplemental sec:octave/octave}

In much of the paper, we have used linearly spaced quantization
levels, which we (and others~\cite{Miyashita2016}) have found is
better matched to the distributions of activation-output values than
is octave-based quantization.  However, if we quantize  our
weights/biases \textbf{and} the activation outputs using uniform log-amplitude
quantization, there are two benefits.  We can:

\begin{itemize}

\item{replace the multiplication (double indexed) LUT of size $N_a
  N_w$  by two much smaller (single-index) tables.  The first smaller
  table is used to move out of log-linear space to linear amplitude
  space, and is of size $max (N_{q;a}, N_{q;w})$.  The second table to
  reverse this mapping is of size $4N_{q;a}$.}

  \item {use only
fixed-point addition to complete the multiplication, by operating in
the $\log_2$ index space.}
\end{itemize}    

Multiplications between quantized weights and quantized activation
outputs become fixed-point additions: using the notation,
\begin{eqnarray*}
  s_x & = & \text{sign}(x) \\
  v_x & = & \text{round}(N_{q,x} \log_2(|x|))
\end{eqnarray*}
where
$N_{q,x}$ is the number of equally spaced samples per octave that we
are using for $x$, then for a $v_{wa}$ quantized to $N_{q;a}$ samples
per octave:
\begin{eqnarray*}
s_{wa} & = & \text{sign}(w a) = s_w s_a \\
v_{wa} & = & \text{round}(N_{q;a} \log_2(|w a|)) \\
 & = & \text{round}(N_{q;a} \log_2(2^{v_w/N_{q;w}} 2^{v_a/N_{q;a}}))
\\
& = & v_a + v_w \frac{N_{q;a}}{N_{q;w}}
\end{eqnarray*}

As long as $N_{q;a}$ is a power of two
 $\times N_{q;w}$, this combination can be done using a bit shift
of $v_w$ followed by a fixed-point addition.  

\begin{figure}
\centering

  \centering
  \begin{small}
  \begin{minipage}{0.65\linewidth}
  \includegraphics[width=\linewidth]{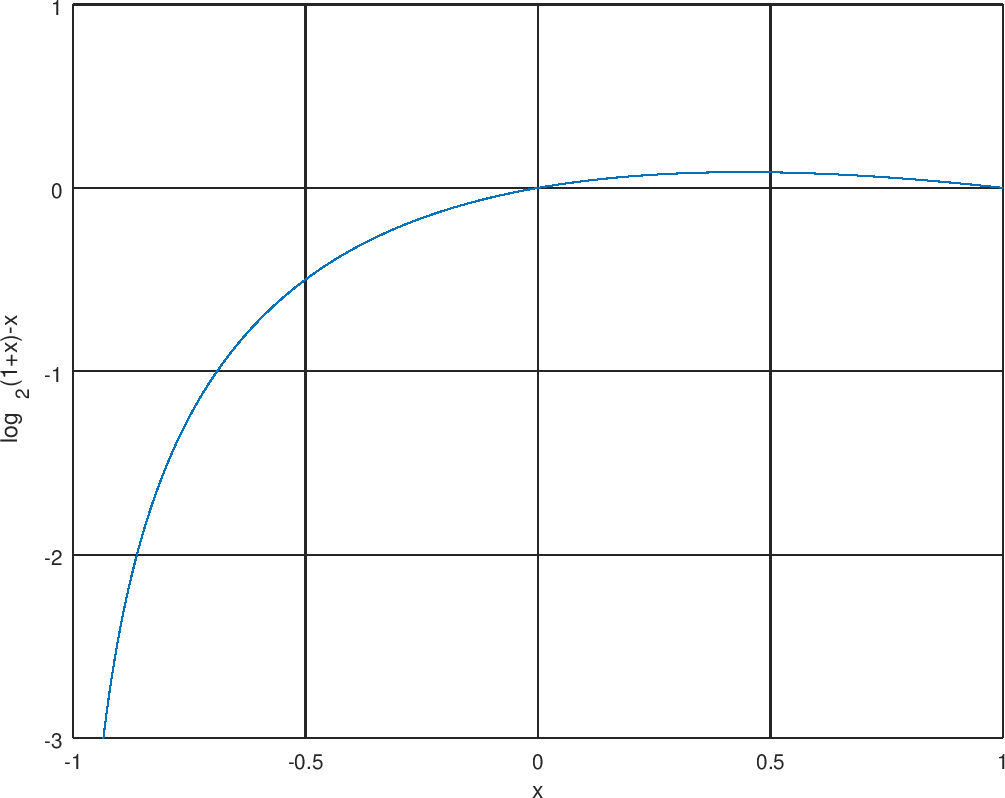}
  \end{minipage}\hfill\begin{minipage}{0.3\linewidth}

\caption{Error in first-order Taylor
  series expansion of $\log_2(1 
  + x)$, expanded around zero, for $-1 < x < 1$}
\label{fig:Taylor}
  \end{minipage}
  \end{small}

\end{figure}

To add together the weighted inputs from the network, we temporarily
move back into (and then out of) the linear amplitude space, using two
singly-indexed LUTs
of combined size $\max (N_{q;a}, N_{q;w}) + 4 N_{q;a}$.
We chose to do this instead of using a
Taylor-series approximation~\cite{Miyashita2016} for accuracy.
Figure~\ref{fig:Taylor} shows the error between
$\log_2(1 + x)$ and its first-order Taylor-series expansion
around $x=0$ for the range from -1 to 1, which is the range needed to use this
expansion as the basis for
linear-value addition in the log domain.  However, as can
be seen by Figure~\ref{fig:Taylor}, that
approximation does not work well when the operands
have opposite signs.  Some of the addends will have opposite signs,
due to negative-signed weights.

We return to the linear amplitude space (before the accumulation)
and map back to the log amplitude space using two small singly-indexed
arrays.  The first table gives us the mapping from
$$i_x = v_x \mod N_{q;a}$$ to\footnote{The actual
  values that are returned from $T_q(i_x)$ are scaled up by 
$N_{o;a} + 3$ (as described below), to allow us to use integer
arithmetic but, for ease and clarity of discussion, we don't include
that power-of-two scaling for the fixed point computation in this
portion.}
$$T_q(i_x) = 2^{i_x/N_{q;a}}$$
As long as $N_{q;a}$ is a power of 2, the
modulo operation is done by simply bit masking the $v_x$ with $N_{q;a}
- 1$, so this translation back to linear amplitude is simply the bit
mask and a table look up.  From this looked-up value, we can
get the actual linear value of the weighted unit input as
$$s_x \text{bitshift}(T_q(i_x), o_x)$$
where $o_x = \lfloor \frac{v_x}{N_{q;a}} \rfloor$ and $s_x$ is
the sign bit for $x$.  Similar to the modulo operation, we can determine $o_x$
using a simple shifting operation (without division), as long as
$N_{q;a}$ is a power of two.

The values for the weighted inputs to the unit are then added
together in linear space, using a fixed-point accumulator.  The
fixed-point scale should be $2^{(N_{o;a} + 3)} / S_{\max}$
where $S_{\max} = 2^{\lceil \log_2 \max(a) \rceil}$
is the next power of two larger than the maximum amplitude of the
activation quantizations and $S_{\max} 2^{-(N_{o;a} + 1)}$ is just
below the minimum amplitude of the
activation quantizations.  The additional 2 bits of precision
will allow us to return to the log-amplitude space with full
resolution of even the lowest octave samples.

If we find that we need to keep precision for the lowest octave when adding and
subtracting %
entries from the highest octave, we can ensure that 
$T_q( )$ (the log-amplitude-to-linear-amplitude table) has entries with
at least $N_{o;a} + 3$ bits. The accumulator itself will need to have
additional ``head room'' beyond
these $(N_{o;a} + 3)$ bits, to avoid underflow and overflow. %
Additionally, we can reduce that expected dynamic range that
the accumulator needs to handle using our training examples to
determine the lowest--maximum-variance order in which to add together
the weighted inputs.

Having accumulated all of the weighted inputs, we need to return to
the log-amplitude domain using a linear--to--log-amplitude
table, $T_q^{-1}( )$. To do this reverse mapping:
\begin{enumerate}
  \item We note the sign of the accumulated value and use that sign as
    our new $s_x$.
  \item We determine the octave of the result, $o_x$, by counting the
    number of leading zeros in our accumulator and adjusting for
    $S_{\max}$ and for the accumulator head room.  This leading-zeros
    count can be done efficiently, without conditional logic, using
    the approach named ``nlz10a()'' in~\cite{Warren2013}.
  \item We use the $N_{o;a} + 2$ bits {\em just below} the leading
    non-zero (lopping off the leading 1) as our index into the
    linear-to-log-amplitude table
    $T_q^{-1}(v) = \log_2(2^{v/(4*N_{q;a})})$.
\end{enumerate}
The final accumulator-output in log representation will
have the sign $s_x$ (from step 1) and the value $-o_x + T_q^{-1}(v_x)$
(from steps 2 and 3).  Using $4 N_{q;a}$ entries in this
reverse-mapping table allows us to distinguish the values at the low end
of the octave.

In summary, we have demonstrated how to eliminate all multiplications
by moving to log-amplitude quantization.  We have found that
this representation is well suited for weights. For representing
activations, empirically, it does not fit as well in the networks we
have tried.  For example, with AlexNet, we needed
more quantization levels when using octave-based sampling on the
activations than we needed using simple linear sampling.  Upon further
analysis, we noted that to achieve the same level of performance we
needed to ensure that the largest step size between activation levels
was no bigger than the step size for the linearly spaced activation
levels to which we are comparing.

\end{document}